\documentclass{article}
\usepackage{ijcai19}

\usepackage{times}
\usepackage{soul}
\usepackage{url}
\usepackage[hidelinks]{hyperref}
\usepackage[utf8]{inputenc}
\usepackage[small]{caption}
\usepackage{graphicx}
\usepackage{epsfig}
\usepackage{epstopdf}
\usepackage{amsmath}
\usepackage{amssymb}
\usepackage{multirow}
\usepackage{booktabs}
\usepackage{algorithm}
\usepackage{algorithmic}
\title{Hierarchical Memory Decoding for Video Captioning}

\author{
Aming Wu\and
Yahong Han
\affiliations
College of Intelligence and Computing, Tianjin University, Tianjin, China
\emails
\{tjwam, yahong\}@tju.edu.cn
}

\begin{document}

\maketitle

\begin{abstract}
  Recent advances of video captioning often employ a recurrent neural network (RNN) as the decoder. However, RNN is prone to diluting long-term information. Recent works have demonstrated memory network (MemNet) has the advantage of storing long-term information. However, as the decoder, it has not been well exploited for video captioning. The reason partially comes from the difficulty of sequence decoding with MemNet. Instead of the common practice, i.e., sequence decoding with RNN, in this paper, we devise a novel memory decoder for video captioning. Concretely, after obtaining representation of each frame through a pre-trained network, we first fuse the visual and lexical information. Then, at each time step, we construct a multi-layer MemNet-based decoder, i.e., in each layer, we employ a memory set to store previous information and an attention mechanism to select the information related to the current input. Thus, this decoder avoids the dilution of long-term information. And the multi-layer architecture is helpful for capturing dependencies between frames and word sequences. Experimental results show that even without the encoding network, our decoder still could obtain competitive performance and outperform the performance of RNN decoder. Furthermore, compared with one-layer RNN decoder, our decoder has fewer parameters.
\end{abstract}

\section{Introduction}

For video captioning, the state of the art methods often follow the encoder-decoder or sequence-to-sequence framework \cite{sutskever2014sequence}. Particularly, in the encoder, they first employ an RNN unit to obtain the representation of the whole video clip. And based on the video representation, in the decoder, they often use a RNN, e.g., LSTM \cite{hochreiter1997long} and GRU \cite{chung2014empirical}, to generate captions. Although LSTM decoder has memory unit to memorize information and obtains good performance of the task \cite{jia2015guiding,vinyals2015show}, as is shown in Weston et al. \cite{memorynetwork}, memorized information in the LSTM cell is limited to several time steps, because the long-term information is gradually diluted at each time step.

Recent methods \cite{gehring2017convolutional,kaiser2017depthwise} explored the utilization of CNN for sequence modeling. As is constrained by the convolutional kernel, the one-layer structure could not model sequences. The CNN model often employs a hierarchical structure, i.e., stacking multiple convolution layers, to model sequences. Although the hierarchical CNN structure has been demonstrated to be effective in tasks like machine translation \cite{gehring2017convolutional}, the performance is constrained by the size of the convolutional kernel and cannot fully capture the information of each layer. Thus, for image captioning, sequence modeling with CNN does not obtain comparable performance \cite{aneja2018convolutional}.

Recent efforts have demonstrated that MemNet \cite{memorynetwork,sukhbaatar2015end} is effective in many tasks, e.g., video question answering \cite{tapaswi2016movieqa} and natural language processing \cite{sukhbaatar2015end}. As MemNet has a mechanism of storing long-term information, it has the ability to capture important elements from the input sequence and alleviate the loss of memory information. Based on the ability, we explore to employ the MemNet-based method to construct the decoder.

\begin{figure*}
\centering
\includegraphics[width=0.95\linewidth]{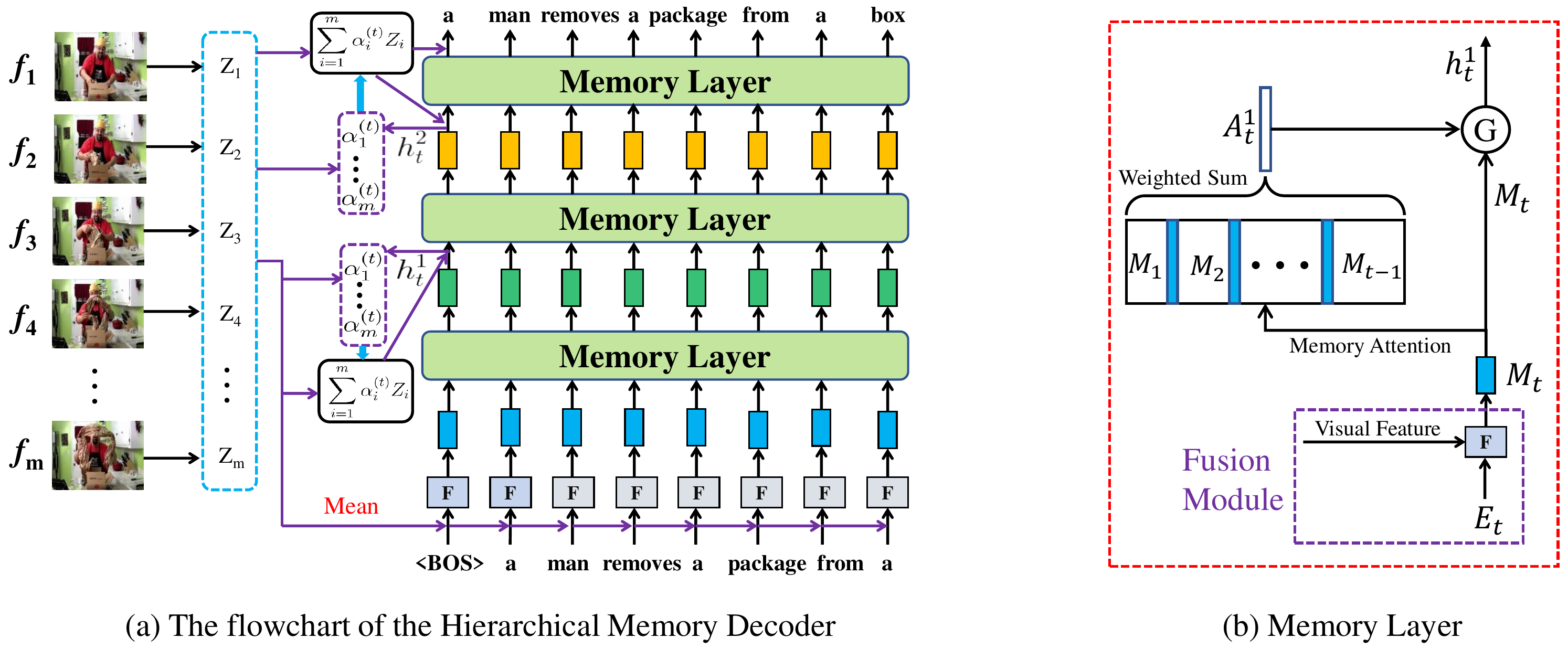}
\caption{(a) is the flowchart of the hierarchical memory decoder. At each time step, we construct a MemNet to generate the word of the next time step. Here we take the decoder consisting of three memory layers as the example. It is worth noting that our network only has the decoder and not have the encoder. (b) indicates the detail of the first memory layer. Here `G' indicates the gated activation.}
\label{introduction}
\end{figure*}

In this paper, we devise a hierarchical memory decoder for video captioning. The framework is shown in Fig. \ref{introduction}. Particularly, after obtaining the representation of each frame, we first fuse the visual and lexical features. Here, we devise a new multi-modal fusion method, i.e., cross-convolution multi-modal fusion (CCMF). Then, at each time step, we construct a MemNet-based decoder. Meanwhile, to fully capture the sequential information, we stack five memory layers as the decoder. And in each layer, we employ the soft-attention mechanism \cite{yao2015describing} to obtain elements which are related to the current input. Finally, in order to reduce the risk of vanishing gradients, we use a multi-layer cross-entropy loss to train this decoder.


In experiments, we evaluate our method on two benchmark datasets of MSVD \cite{chen2011collecting} and MSRVTT \cite{xu2016msr}. To the best of our knowledge, ours is the first memory sequence decoder for video captioning. And compared with several baseline methods which use a single kind of visual feature as input, our method could obtain competitive performance and outperforms the performance of RNN decoder.

\section{Related Work}

The goal of video captioning is to generate a sentence to describe the video content. The state of the art methods often follow encoder-decoder framework \cite{venugopalan2015sequence}. Thus, most methods improve the encoder or the decoder to improve the performance.

\begin{figure}
\centering
\includegraphics[width=0.95\linewidth]{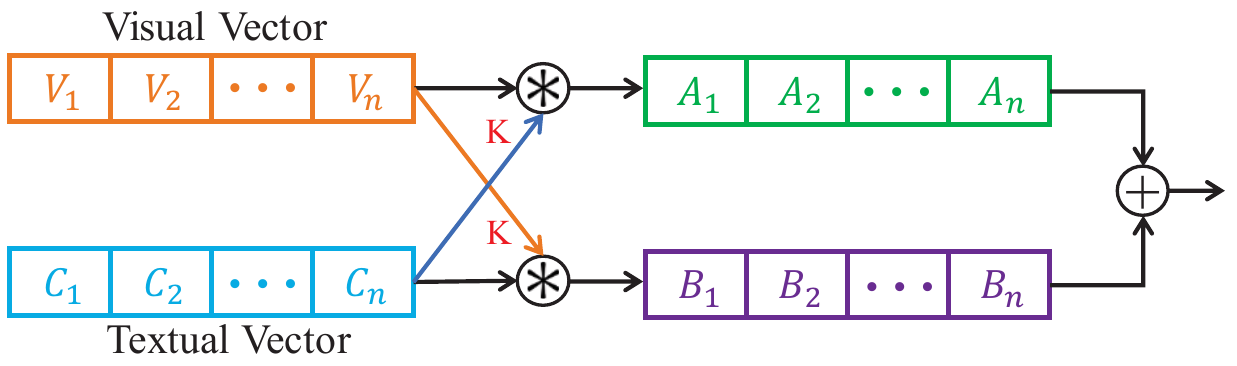}
\caption{Cross-Convolution Multi-modal Fusion. Here `K' indicates the convolutional kernel. $\circledast$ indicates the convolution operation. $\oplus$ indicates the element-wise sum.}
\label{fusion}
\end{figure}

Particularly, for the encoder, Pan et al. \cite{pan2016hierarchical} proposed a hierarchical recurrent neural encoder to fully capture the sequential information of the input video. And Baraldi et al. \cite{baraldi2017hierarchical} proposed a hierarchical boundary-aware encoder for video captioning. This encoder could fully capture the discriminative feature among video frames. Besides, Wang et al. \cite{wang2018m3} designed a memory network-based framework to leverage much more video content. The work \cite{chen2018less} designed a method to pick informative frames to generate caption. In short, the role of the encoder is to fully leverage video content to improve the quality of the generated caption.

Compared with encoder, the work about the improvement of decoder is relatively less. The commonly used decoder is still LSTM-based framework \cite{hochreiter1997long}. The work \cite{yao2015describing} proposed an attention mechanism to help the decoder capture the video content which is related to the generated word. Song et al. \cite{song2017hierarchical} proposed a hierarchical LSTM as the decoder. However, as using multiple LSTM units in the decoder, this method increased the number of parameters and computational costs. Recently, Mehri et al. \cite{mehri2018middle} proposed a middle-out decoding method to improve the LSTM. Though this method improves the decoding efficiency, it still could not solve the problem of long-term information dilution. In this paper, we propose a MemNet-based hierarchical decoder. As the MemNet has the advantage of storing long-term information, our decoder avoids losing long-term information. Moreover, employing the hierarchical structure and multi-layer loss could not only promote the decoder to model sequences and reduce the risk of vanishing gradients but also help the decoder gradually learn to generate accurate captions. Experimental results on two benchmark datasets demonstrate the effectiveness of our decoder.

\section{Hierarchical Memory Decoder}

In this section, we delve into the main contribution of this paper, i.e., a hierarchical memory decoder with a new multi-modal fusion method.

\textbf{Feature Extraction.} Given a video including $m$ frames, we first employ a pre-trained deep convolutional neural network \cite{szegedy2015going,he2016deep} to extract feature for each of the $m$ frames, which results in a vector $X_{i}\in\mathbb{R}^{q}$ for the $i$-th frame. In order to reduce parameters and computational cost, we employ a filter $W_{c} \in \mathbb{R}^{1 \times q \times n}$ ($n \le q$) to obtain lower dimensional representation $Z_{i} \in \mathbb{R}^{n}$ of $X_{i}$. Meanwhile, we use $V \in \mathbb{R}^{n}$ to indicate the mean of these lower dimensional representations.

\subsection{Cross-Convolution Multi-modal Fusion}

As shown in Fig. \ref{introduction}(a), the input of our hierarchical memory decoder includes the visual and lexical information. Thus, we should first fuse these two multi-modal information. The common used multi-modal fusion methods mainly include element-wise addition, element-wise product, concatenation, bilinear pooling \cite{fukui2016multimodal} and circulant fusion \cite{MCF}. And recent work \cite{fukui2016multimodal} has shown that bilinear pooling is an effective fusion method. This inspires us that exploiting complex interactions among the feature dimensions is helpful for capturing the common semantics of multi-modal features.

To better exploit complex interaction and reduce the number of parameters, we design a CCMF method (Fig. \ref{fusion}). Given two feature vectors in different modalities, e.g., the visual feature $V \in \mathbb{R}^{n}$ and the lexical feature $C \in \mathbb{R}^{n}$. We respectively take the visual feature and the lexical feature as the 1-dimensional convolution kernel to make 1-dimensional convolution operation. The details are shown as follows:

\begin{equation}\label{fusion:eq1}
\begin{split}
& kernel1 = VW_{1}^{T} \\
& kernel2 = CW_{2}^{T} \\
& A = kernel2 \circledast V \\
& B = kernel1 \circledast C \\
& M = ReLU(A) + ReLU(B)
\end{split}
\end{equation}

\noindent where $W_{1}^{T}$ and $W_{2}^{T}$ are the transpose of $W_{1} \in \mathbb{R}^{n \times n}$ and $W_{2} \in \mathbb{R}^{n \times n}$. $M$ is the final fusion result. Here, for the convenience of calculation, we set the dimensions of the visual feature and lexical feature the same. The goal of convolution operation is to make the multi-modal elements fully interact.

\subsection{Hierarchical Memory Decoder Architecture}

In this paper, we stack five identical memory layers to form the memory sequence decoder. In the following, we denote the predicted word sequence by $\hat{Y} = \{\hat{Y}_{1}, ... , \hat{Y}_{T}\}$. We denote the target word sequence by $Y = \{Y_{1}, ..., Y_{T}\}$, where $T$ denotes sequence length. `$\odot$' is an element-wise product operation. $Z = \{Z_{1}, Z_{2}, \cdots, Z_{m}\}$. $\sigma(\cdot)$ is a sigmoid function. In this paper, we use gated activation unit \cite{oord2016pixel}. $attention(Q, S)$ represents the attention operation based on the query vector $Q$ and the feature set $S$.

\textbf{First Memory Layer.} At each time step $t$, we take the fusion result $M_{t}$ as the input of the first layer. The operations of this layer are shown as follows:

\begin{equation}\label{decoder:eq1}
\begin{split}
& S_{t-1}^{1} = [M_{1}, M_{2}, \cdots, M_{t-1}] \\
& A_{t}^{1} = attention(M_{t}, S_{t-1}^{1}) \\
& h_{t}^{1} = tanh(w_{f}^{1} \circledast M_{t} + b_{f}^{1}) \odot \sigma(w_{g}^{1} \circledast A_{t}^{1} + b_{g}^{1})
\end{split}
\end{equation}

\noindent where $M_{t}$ represents the fusion result. $S_{t-1}^{1}$ is the set of fusion result from the time step 1 to $t-1$. $w_{f}^{1}$ and $w_{g}^{1}$ denote convolutional filters on the first layer, which are used to adjust the number of channels of $M_{t}$ and $A_{t}^{1}$. $b_{f}^{1}$ and $b_{g}^{1}$ denote bias on the first layer.

\textbf{Second Memory Layer.} For the second layer, we first use the output $h_{t}^{1}$ of the first layer to compute visual attention $\varphi_{t}^{1}(Z)$. Then we take the concatenation of $\varphi_{t}^{1}(Z)$ and $h_{t}^{1}$ as the input. The operations of this layer are shown as follows:

\begin{equation}\label{decoder:eq2}
\begin{split}
& I_{t}^{2} = w_{2} \circledast |h_{t}^{1}, \varphi_{t}^{1}{(Z)}| + b_{2} \\
& S_{t-1}^{2} = [I_{1}^{2}, I_{2}^{2}, \cdots, I_{t-1}^{2}] \\
& A_{t}^{2} = attention(I_{t}^{2}, S_{t-1}^{2}) \\
& h_{t}^{2} = tanh(w_{f}^{2} \circledast I_{t}^{2} + b_{f}^{2}) \odot \sigma(w_{g}^{2} \circledast A_{t}^{2} + b_{g}^{2})
\end{split}
\end{equation}

\noindent where $w_{2}$ is a learnable filter to convert the channel of concatenated representation. $|a, b|$ represents the concatenation of $a$ and $b$. $w_{f}^{2}$ and $w_{g}^{2}$ denote convolutional filters on the second layer, which are used to adjust the number of channels of $I_{t}^{2}$ and $A_{t}^{2}$. $b_{f}^{2}$ and $b_{g}^{2}$ denote bias on the second layer.

Then, the operations of the next two layers are as:

\begin{equation}\label{decoder:eq3}
\begin{split}
& S_{t-1}^{l} = [h_{1}^{l-1}, h_{2}^{l-1}, \cdots, h_{t-1}^{l-1}] \\
& A_{t}^{l} = attention(h_{t}^{l-1}, S_{t-1}^{l}) \\
& h_{t}^{l} = tanh(w_{f}^{l} \circledast h_{t}^{l-1} + b_{f}^{l}) \odot \sigma(w_{g}^{l} \circledast A_{t}^{l} + b_{g}^{l})
\end{split}
\end{equation}

\noindent where $h_{t}^{l}$ represents the output of $l$-th layer at the time step $t$. $S_{t-1}^{l}$ is the set of the output of $(l-1)$-th layer from the time step 1 to $t-1$. $w_{f}^{l}$ and $w_{g}^{l}$ denote convolutional filters on the $l$-th layer, which are used to adjust the number of channels of $h_{t}^{l-1}$ and $A_{t}^{l}$. $b_{f}^{l}$ and $b_{g}^{l}$ denote the bias on the $l$-th layer.

\textbf{Output Layer.} For the output layer, we first use the output $h_{t}^{4}$ of the fourth layer to compute visual attention $\varphi_{t}^{4}(Z)$. Then we take the sum of $h_{t}^{4}$ and $\varphi_{t}^{4}(Z)$ as the input. The operations of this layer are shown as follows:

\begin{equation}\label{decoder:eq4}
\begin{split}
& I_{t}^{5} = h_{t}^{4} + \varphi_{t}^{4}{(Z)} \\
& S_{t-1}^{5} = [I_{1}^{5}, I_{2}^{5}, \cdots, I_{t-1}^{5}] \\
& A_{t}^{5} = attention(I_{t}^{5}, S_{t-1}^{5}) \\
& h_{t}^{5} = tanh(w_{f}^{5} \circledast I_{t}^{5} + b_{f}^{5}) \odot \sigma(w_{g}^{5} \circledast A_{t}^{5} + b_{g}^{5})
\end{split}
\end{equation}

\noindent where $w_{f}^{5}$ and $w_{g}^{5}$ denote convolutional filters on the 5th layer, which are used to adjust the number of channels of $I_{t}^{5}$ and $A_{t}^{5}$. $b_{f}^{5}$ and $b_{g}^{5}$ denote the bias.

Finally, the $t$-th generated word $\hat{Y}_{t}$ is computed as follows:

\begin{equation}\label{decoder:eq5}
\begin{split}
&\hat{Y}_{t} \sim softmax(w_{p}(h_{t}^{5} + \varphi_{t}^{4}{(Z)}) + b_{p})
\end{split}
\end{equation}

\noindent where $w_{p}$ and $b_{p}$ are learnable projection matrix and bias.

\textbf{Cold-start Processing.} For video caption generation, there is no future information available for the decoder. Besides, for the time step 1, as there is no previous information using for attention computation, we need to make some special process for this step. Particularly, we generate a random vector $H$ which is from a normal distribution and has the same shape as $V$. Then we take the sum of $H$ and $V$ as the input of the first layer. And for the next four layers, we all generate a random vector for each layer and take the sum of the vector and $h_{1}^{l}$ as the input of next layer.

\textbf{Attention Mechanism.} In the decoder, there are two types of attention, i.e., visual attention and memory attention (as shown in Fig. \ref{introduction}(b)). And for these two types of attention, we all use soft attention mechanism \cite{yao2015describing}. Concretely, for memory attention, at step $t$, based on the set $S_{t-1}^{l}$, we first compute the dynamic attention weight $\alpha_{i}^{(t)}$.

\begin{equation}\label{decoder:eq6}
\begin{split}
& e_{i}^{(t)} = w^{T}tanh(W_{a}h_{t}^{l-1} + U_{a}S_{t-1}^{l}[i] + b_{a}) \\
& \alpha_{i}^{(t)} = exp{\{e_{i}^{(t)}\}} / \sum_{j=1}^{t-1}exp{\{e_{j}^{(t)}\}}
\end{split}
\end{equation}

\noindent where $w$, $W_{a}$, $U_{a}$, and $b_{a}$ are learnable parameters. $w^{T}$ is the transpose of $w$. $h_{t}^{l-1}$ represents the output of $(l-1)$-th layer.

Finally, we take the dynamically weighted sum of the set $S_{t-1}^{l}$ as the attention result of the $l$-th layer.

\begin{equation}\label{attend:eq2}
\varphi_{t}^{l}(S) = \sum_{i=1}^{t-1}\alpha_{i}^{(t)}S_{t-1}^{l}[i]
\end{equation}

And the process of visual attention is same as that of memory attention. Here we replace the set $S_{t-1}^{l}$ with $Z = (Z_{1}, Z_{2}, ...,Z_{m})$.

\textbf{Training Loss.} In order to reduce the risk of vanishing gradients, inspired by the work \cite{zhang2016augmenting}, we enforce intermediate supervision for some hidden layers. In this paper, we empirically enforce supervision for the first, third, and fifth memory layer. For each layer $j\in\{1,3,5\}$, we employ a cross-entropy loss.

\begin{equation}\label{loss}
L_{XE}^{j} = -\sum_{t=1}^{T} log(p(Y_{t}|Y_{1:t-1}, Z))
\end{equation}

\noindent where $Y_{t}$ is the ground-truth word at time $t$, $p(Y_{t}|Y_{1:t-1}, Z)$ is the output probability of word $Y_{t}$ given the previous word $Y_{1:t-1}$ and encoding output $Z$. By summing the loss of each layer, we obtain the training loss of memory decoder:

\begin{equation}\label{all_loss}
L_{XE} = \lambda_{1}L_{XE}^{1} + \lambda_{3}L_{XE}^{3} + \lambda_{5}L_{XE}^{5}
\end{equation}

\noindent where $\lambda_{1}$, $\lambda_{3}$, and $\lambda_{5}$ are hyper-parameters. Here we keep the sum of $\lambda_{1}$, $\lambda_{3}$, and $\lambda_{5}$ is 1. Besides, as the fifth layer is the output layer, we should keep $\lambda_{5}$ is larger than $\lambda_{1}$ and $\lambda_{3}$.

\section{Experiments}

In the following, we first compare our method with some baseline methods. Then we make some ablation analysis about our method. All results are evaluated by metrics of BLEU \cite{papineni2002bleu}, METEOR \cite{denkowski2014meteor}, and CIDEr \cite{vedantam2015cider}.

\subsection{Dataset and Implementation Details}

\textbf{Datasets.} MSVD \cite{chen2011collecting} contains 1,970 video clips with multiple descriptions for each video clip. Following the work \cite{venugopalan2014translating}, we use 1,200 video clips for training, 100 video clips for validation, and 670 video clips for testing. MSRVTT \cite{xu2016msr} is the largest dataset for video captioning. The dataset contains 10,000 video clips. Following the work \cite{xu2016msr}, we use 6,513 videos for training, 497 videos for validation, and 2,990 videos for testing.

\textbf{Video Processing.} For the MSVD dataset, we select 40 equally-spaced frames from each video and feed them into GoogLeNet \cite{szegedy2015going} and ResNet-152 \cite{he2016deep} to extract a 1,024 and 2,048-dimensional frame-wise representation. For the MSRVTT dataset, we select 20 equally-spaced frames from each video and feed them into GoogLeNet and ResNet-152 to extract 1,024 and 2,048-dimensional representation, respectively.

\textbf{Parameters Setting.} In the experiment, we set the channel $n$ of $Z_{i}$ to 512. For each memory layer, we set $w \in \mathbb{R}^{100 \times 1}$, $W_{a} \in \mathbb{R}^{100 \times 512}$, $U_{a} \in \mathbb{R}^{100 \times 512}$, $b_{a} \in \mathbb{R}^{100}$ (Eq. \eqref{decoder:eq6}), and $w_{f}^{l} \in \mathbb{R}^{1 \times 512 \times 512}$, $w_{g}^{l} \in \mathbb{R}^{1 \times 512 \times 512}$, $b_{f}^{l} \in \mathbb{R}^{512}$, $b_{g}^{l} \in \mathbb{R}^{512}$, $l \in \{1,\cdots,5\}$, (Eq. \eqref{decoder:eq3}). The parameter settings of visual attention are same as those of the memory layers.

\textbf{Training Details.} The vocabulary size is 12,596 for MSVD and 23,308 for MSRVTT, respectively. During training, all parameters are randomly initialized. We use Adam optimizer with an initial learning rate of $1\times10^{-3}$ and a momentum parameter of 0.9. We empirically set $\lambda_{1}$, $\lambda_{3}$, and $\lambda_{5}$ to 0.2, 0.2, and 0.6, respectively. Note that we do not conduct beam search in testing.

\subsection{Experiment Results}

As our method is only a decoder with attention mechanism, here for fair comparison, we compare some baseline methods, e.g., the work \cite{yao2015describing} and the work \cite{song2017hierarchical}. And these methods include different types of the decoder and rarely encode the video content.

\textbf{MSVD Dataset.} On MSVD dataset, we compare our method with some baseline methods. The results are shown in Table \ref{MSVD}. Particularly, compared with the work \cite{yao2015describing} which only includes LSTM decoder and visual attention, our method outperforms its performance obviously. This shows that our decoder is very effective. Although the methods \cite{ballas2015delving,pan2016jointly,yu2016video} use different types of RNN as the decoder, the performance of our method also outperforms them. Moreover, the work \cite{song2017hierarchical} stacks multiple LSTM layers as the decoder. However, its performance is also weaker than us on the metric of `METEOR'. Besides, stacking multiple LSTM layers often leads to the increase of training time and computational cost. These all demonstrate our decoder could obtain comparable performance as RNN decoder. Meanwhile, the amount of parameter in our decoder is 4.13 M, which is fewer than that of one-layer LSTM whose amount of parameter is 4.28 M. This shows our hierarchical memory decoder is effective. Besides, the Middle-out \cite{mehri2018middle} is a newly proposed decoder. Compared with Middle-out decoder, our method still outperforms it. And we do not use encoding network. This further shows the effectiveness of our decoder.

\begin{table}[ht]
  \begin{center}
  \scriptsize
   \caption{Comparison with other models. Here `G' denotes GoogLeNet. `R' denotes ResNet. `Memory' represents our memory sequence decoder. `self' denotes self-attention. All values are measured by percentage (\%).}\label{MSVD}
  \begin{tabular}{c|c|c|c}
  \hline
  Method &BLEU@4 &METEOR &CIDEr \\
  \hline
  S2VT \cite{venugopalan2015sequence} & - & 29.20 & - \\
  VGG+LSTM-E \cite{pan2016jointly} & 40.20 & 29.50 & - \\
  C3D+LSTM-E \cite{pan2016jointly} & 41.70 & 29.90 & - \\
  VGG+p-RNN \cite{yu2016video} & 44.30 & 31.10 & 62.10 \\
  C3D+p-RNN \cite{yu2016video} & 47.40 & 30.30 & 53.60 \\
  Tempor-attention \cite{yao2015describing} & 41.92 & 29.60 & 51.67 \\
  G+Bi-GRU-RC$\rm N_{1}$ \cite{ballas2015delving} & 48.42 & 31.70 & 65.38 \\
  G+Bi-GRU-RC$\rm N_{2}$ \cite{ballas2015delving} & 43.26 & 31.60 & 68.01 \\
  G+hLSTMat \cite{song2017hierarchical} & 48.50 & 31.90 & - \\
  C3D+hLSTMat \cite{song2017hierarchical} & 47.50 & 30.50 & - \\
  MAMRNN \cite{li2017mam} & 41.40 & 32.20 & 53.90 \\
  R+PickNet \cite{chen2018less} & 46.10 & 33.10 & 76.00 \\
  MCF \cite{MCF} & 46.46 & 33.72 & 75.46 \\
  RecNet \cite{wang2018reconstruction} & \textbf{51.10} & 34.10 & 80.30 \\
  Middle-out \cite{mehri2018middle} & 40.80 & 30.90 & 68.60 \\
  Middle-out+self \cite{mehri2018middle} & 47.00 & 34.10 & 79.50 \\
  \hline
  G+Memory & 45.74 & 33.01 & 73.11 \\
  R+Memory & 49.28 & \textbf{34.67} & \textbf{81.49} \\
  \hline
  \end{tabular}
  \end{center}
\end{table}

\begin{figure*}
\centering
\includegraphics[width=0.95\linewidth]{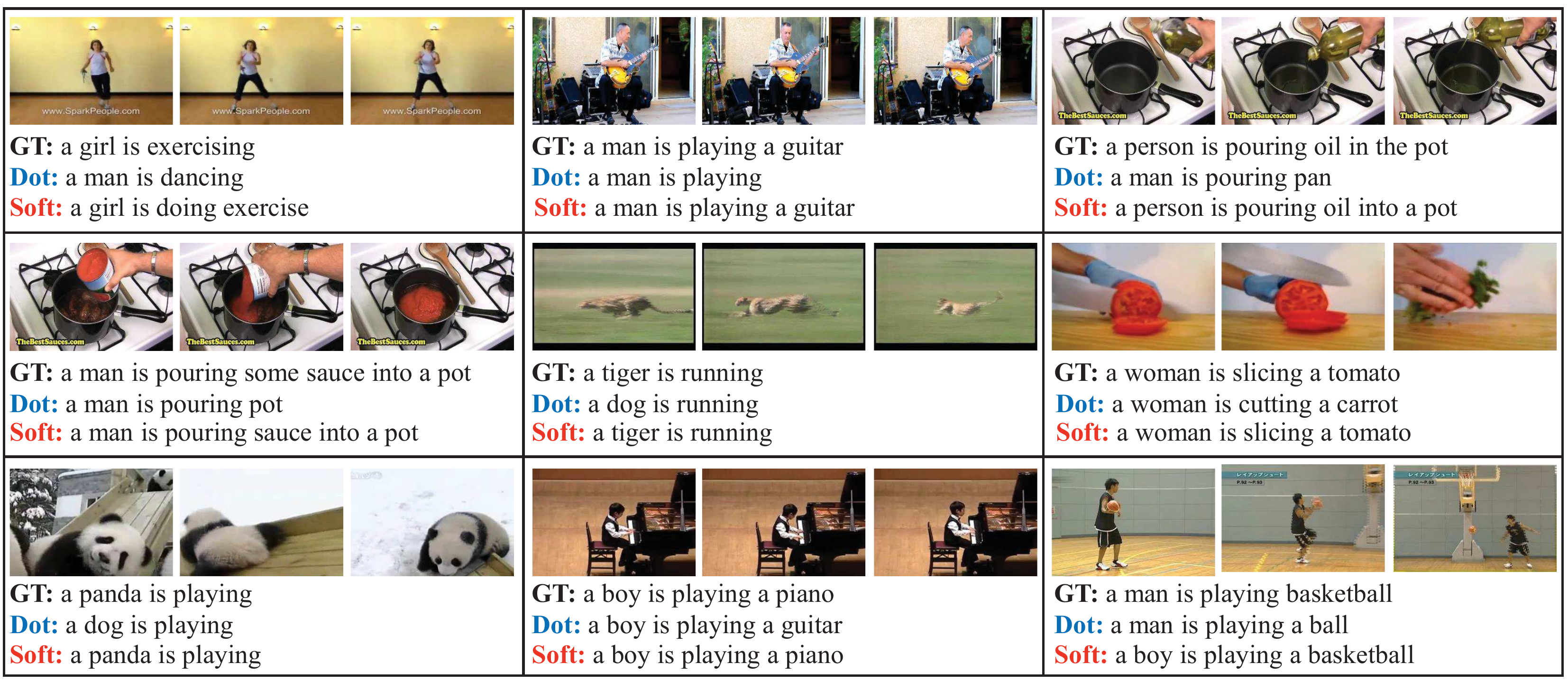}
\caption{Examples of dot-product attention and soft attention. Here, `GT' represents the ground truth. `Dot' represents using dot-product attention in our decoder. `Soft' represents the soft attention.}
\label{attention}
\end{figure*}

\textbf{MSRVTT Dataset.} On MSRVTT dataset, we compare our method with some state of the art methods. Compared with the work \cite{song2017hierarchical}, our method also outperforms its performance on the metric of `METEOR'. Compared with the work \cite{venugopalan2014translating} which uses the GoogLeNet feature as input, our method obviously outperforms its performance. For the work \cite{xu2017learning} which uses multiple kinds of features as input, our method also outperforms its performance on the three metrics. Besides, the work \cite{phan2017consensus} uses the reinforcement learning (RL) method for video captioning, which is much more complex than our method. However, we can see that our method is similar to its performance. These all show that our method is effective.

\begin{table}[ht]
  \begin{center}
  \scriptsize
  \caption{Comparison with other models on MSRVTT. Note that we use the dataset of MSRVTT 2016. Here `G' and `R' denote GoogLeNet and ResNet features. `Memory' represents our hierarchical memory decoder. All values are measured by percentage (\%).}\label{MSRVTT}
  \begin{tabular}{c|c|c|c}
  \hline
  Method &BLEU@4 &METEOR &CIDEr \\
  \hline
  MA-LSTM \cite{xu2017learning} & 36.5 & 26.5 & 41.0 \\
  G+LSTM \cite{venugopalan2014translating} & 34.6 & 24.6 & - \\
  C3D+SA \cite{yao2015describing} & 36.1 & 25.7 & - \\
  R+S2VT \cite{sutskever2014sequence} & 31.4 & 25.7 & 35.2 \\
  R+hLSTMat \cite{song2017hierarchical} & \textbf{38.3} & 26.3 & - \\
  R+Consensus \cite{phan2017consensus} & 37.5 & 26.6 & 41.5 \\
  M3-VC \cite{wang2018m3} & 38.1 & 26.6 & - \\
  \hline
  G+Memory & 35.7 & 26.1 & 37.8 \\
  R+Memory & 37.5 & \textbf{26.9} & \textbf{41.7} \\
  \hline
  \end{tabular}
  \end{center}
\end{table}

\subsection{Ablation Analysis}

In the following, we make some ablation analysis about our method on the MSVD dataset. We use GoogLeNet feature.

\textbf{Fusion Method.} In order to make the decoder fully leverage the video features, at each time step, the input of the decoder includes the lexical feature and the mean representation of video features (Fig. \ref{introduction}(a)). This requires us to use a proper method to fuse the visual and lexical feature. To demonstrate CCMF is effective for our decoder, we compare our method with two effective fusion method, i.e., bilinear pooling \cite{fukui2016multimodal} and circulant fusion \cite{MCF}. We can see from Table \ref{compare_attention} that for our decoder, our fusion method outperforms the compared methods. This shows our fusion method is effective for our decoder.

\begin{table}[ht]
  \begin{center}
  \small
  \caption{Comparison of different fusion methods. All values are measured by percentage(\%).}\label{compare_attention}
  \begin{tabular}{c|c|c|c}
  \hline
  Method &BLEU@4 &METEOR &CIDEr \\
  \hline
  Bilinear Pooling & 45.51 & 32.20 & 70.15 \\
  Circulant Fusion & 44.32 & 32.84 & 72.16 \\
  CCMF & 45.74 & 33.01 & 73.11 \\
  \hline
  \end{tabular}
  \end{center}
\end{table}

\begin{figure*}
\centering
\includegraphics[width=0.95\linewidth]{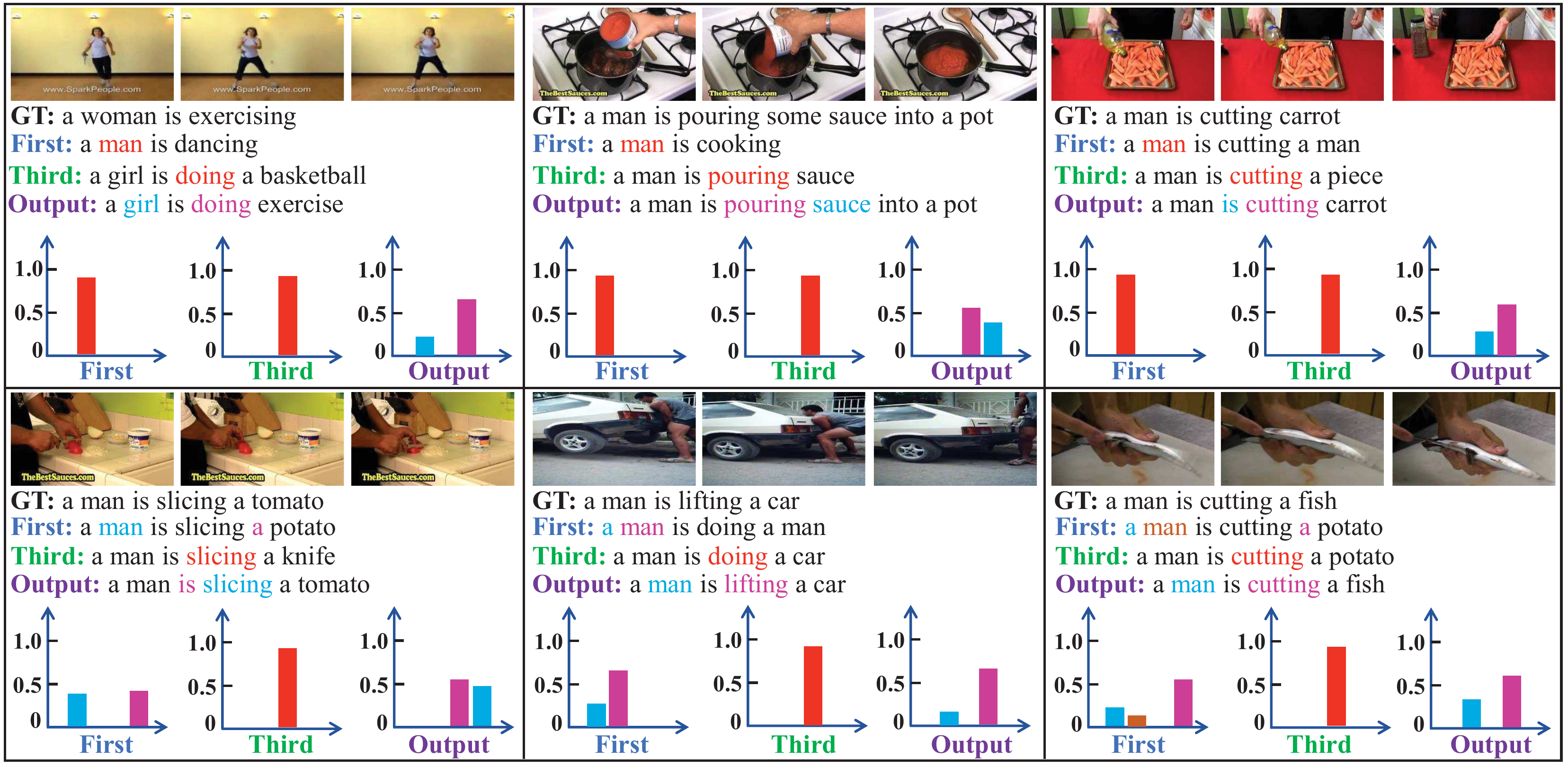}
\caption{Comparison of different memory layer. `First', `Third', and `Output' respectively represents the first, the third, and the output memory layer in our decoder. At the bottom of each example, we plot the attention weight $\alpha_{i}^{(t)}$ for each layer when the last word is generated. The vertical axis represents attention weight. For each layer, the color of the weight corresponds to the color of the word in the sentence. The red color represents the weight $\alpha_{i}^{(t)} \ge 0.95$. And from left to right is the distribution of the attention weights of the first layer, the third layer, and the output layer.}
\label{layer}
\end{figure*}

\textbf{Attention Mechanism.} Dot-product attention \cite{vaswani2017attention} and soft attention \cite{yao2015describing} are the two commonly used attention mechanisms. In order to demonstrate the soft attention used in our decoder is effective, here we replace the soft attention with the dot-product attention and keep other components of the decoder invariant. The results are shown in Table \ref{attention_method}. We can see that in our decoder, the performance of soft attention outperforms dot-product attention. This shows that for our decoder, the soft attention mechanism is effective.

\begin{table}[ht]
  \begin{center}
  \small
  \caption{Comparison of different attention mechanism. `Dot-Attention' denotes the dot product attention. All values are measured by percentage (\%).}\label{attention_method}
  \begin{tabular}{c|c|c|c}
  \hline
  Method &BLEU@4 &METEOR &CIDEr \\
  \hline
  Dot-Attention & 45.95 & 32.24 & 68.62 \\
  Soft-Attention & 45.74 & 33.01 & 73.11 \\
  \hline
  \end{tabular}
  \end{center}
\end{table}

In Fig. \ref{attention}, we show some examples of dot-product attention and soft attention. We can see that the results generated by the soft attention are better than those of dot-product attention. Taking the fifth and seventh results as examples, our method successfully recognizes the `tiger' and `panda'. This further shows that the soft attention mechanism is effective.

\textbf{Performance of Each Memory Layer.} In the following, we analyze the performance of different memory layer. Here, we choose the first, the third, and the output layer (Eq. \eqref{all_loss}). The results are shown in Table \ref{comparison_layer}. We can see that the results from the first memory layer to the output memory layer continuously improve. This shows that stacking multiple memory layers is helpful for capturing the sequence information and improving performance. In order to further show stacking multiple layers is helpful, we show some examples in Fig. \ref{layer}. We can see that from the first layer to the output layer, the accuracy of the generated caption is increasingly getting better. Besides, from the bottom of each example, we can see that as the layer getting deeper, the attention weight of each memory layer changes towards the more important word. Taking the first and the second result as example, when generating the last word `exercise' and `pot', the main attention of output layer is `girl', `doing', and `pouring', `sauce', which are important for the generating of the last word. These all show that the hierarchical memory decoder could capture the relevance corresponding to a generated word. The relevance is helpful for capturing much more important memory and generating accuracy captions.

\begin{table}[ht]
  \begin{center}
  \small
  \caption{Comparison of different memory layer. All values are measured by percentage (\%).} \label{comparison_layer}
  \begin{tabular}{c|c|c|c}
  \hline
  Method &BLEU@4 &METEOR &CIDEr \\
  \hline
  First-Layer & 39.21 & 28.75 & 44.80 \\
  Third-Layer & 45.22 & 32.64 & 70.18 \\
  Output-Layer & 45.74 & 33.01 & 73.11 \\
  \hline
  \end{tabular}
  \end{center}
\end{table}

\section{Conclusion}

In this paper, we propose a new hierarchical memory decoder for video captioning. We first design a new multi-modal fusion mechanism, i.e., CCMF, to fuse visual and lexical information. Then, at each time step, taking the fusion result as the input, we construct a memory network-based decoder to generate captions. Experimental results on two benchmark datasets demonstrate the effectiveness of our method.

\scriptsize
\bibliographystyle{named}
\bibliography{ijcai19}

\begin{thebibliography}{}

\bibitem[\protect\citeauthoryear{Aneja \bgroup \em et al.\egroup
  }{2018}]{aneja2018convolutional}
Jyoti Aneja, Aditya Deshpande, and Alexander~G Schwing.
\newblock Convolutional image captioning.
\newblock In {\em Proceedings of the IEEE Conference on Computer Vision and
  Pattern Recognition}, pages 5561--5570, 2018.

\bibitem[\protect\citeauthoryear{Ballas \bgroup \em et al.\egroup
  }{2015}]{ballas2015delving}
Nicolas Ballas, Li~Yao, Chris Pal, and Aaron Courville.
\newblock Delving deeper into convolutional networks for learning video
  representations.
\newblock {\em arXiv preprint arXiv:1511.06432}, 2015.

\bibitem[\protect\citeauthoryear{Baraldi \bgroup \em et al.\egroup
  }{2017}]{baraldi2017hierarchical}
Lorenzo Baraldi, Costantino Grana, and Rita Cucchiara.
\newblock Hierarchical boundary-aware neural encoder for video captioning.
\newblock In {\em Computer Vision and Pattern Recognition (CVPR), 2017 IEEE
  Conference on}, pages 3185--3194. IEEE, 2017.

\bibitem[\protect\citeauthoryear{Chen and Dolan}{2011}]{chen2011collecting}
David~L Chen and William~B Dolan.
\newblock Collecting highly parallel data for paraphrase evaluation.
\newblock In {\em Proceedings of the 49th Annual Meeting of the Association for
  Computational Linguistics: Human Language Technologies-Volume 1}, pages
  190--200. Association for Computational Linguistics, 2011.

\bibitem[\protect\citeauthoryear{Chen \bgroup \em et al.\egroup
  }{2018}]{chen2018less}
Yangyu Chen, Shuhui Wang, Weigang Zhang, and Qingming Huang.
\newblock Less is more: Picking informative frames for video captioning.
\newblock {\em arXiv preprint arXiv:1803.01457}, 2018.

\bibitem[\protect\citeauthoryear{Chung \bgroup \em et al.\egroup
  }{2014}]{chung2014empirical}
Junyoung Chung, Caglar Gulcehre, KyungHyun Cho, and Yoshua Bengio.
\newblock Empirical evaluation of gated recurrent neural networks on sequence
  modeling.
\newblock {\em arXiv preprint arXiv:1412.3555}, 2014.

\bibitem[\protect\citeauthoryear{Denkowski and
  Lavie}{2014}]{denkowski2014meteor}
Michael Denkowski and Alon Lavie.
\newblock Meteor universal: Language specific translation evaluation for any
  target language.
\newblock In {\em Proceedings of the ninth workshop on statistical machine
  translation}, pages 376--380, 2014.

\bibitem[\protect\citeauthoryear{Fukui \bgroup \em et al.\egroup
  }{2016}]{fukui2016multimodal}
Akira Fukui, Dong~Huk Park, Daylen Yang, Anna Rohrbach, Trevor Darrell, and
  Marcus Rohrbach.
\newblock Multimodal compact bilinear pooling for visual question answering and
  visual grounding.
\newblock {\em arXiv preprint arXiv:1606.01847}, 2016.

\bibitem[\protect\citeauthoryear{Gehring \bgroup \em et al.\egroup
  }{2017}]{gehring2017convolutional}
Jonas Gehring, Michael Auli, David Grangier, Denis Yarats, and Yann~N Dauphin.
\newblock Convolutional sequence to sequence learning.
\newblock {\em arXiv preprint arXiv:1705.03122}, 2017.

\bibitem[\protect\citeauthoryear{He \bgroup \em et al.\egroup
  }{2016}]{he2016deep}
Kaiming He, Xiangyu Zhang, Shaoqing Ren, and Jian Sun.
\newblock Deep residual learning for image recognition.
\newblock In {\em Proceedings of the IEEE conference on computer vision and
  pattern recognition}, pages 770--778, 2016.

\bibitem[\protect\citeauthoryear{Hochreiter and
  Schmidhuber}{}]{hochreiter1997long}
Sepp Hochreiter and J{\"u}rgen Schmidhuber.
\newblock Long short-term memory.
\newblock {\em Neural computation}.

\bibitem[\protect\citeauthoryear{Jia \bgroup \em et al.\egroup
  }{2015}]{jia2015guiding}
Xu~Jia, Efstratios Gavves, Basura Fernando, and Tinne Tuytelaars.
\newblock Guiding long-short term memory for image caption generation.
\newblock {\em arXiv preprint arXiv:1509.04942}, 2015.

\bibitem[\protect\citeauthoryear{Kaiser \bgroup \em et al.\egroup
  }{2017}]{kaiser2017depthwise}
Lukasz Kaiser, Aidan~N Gomez, and Francois Chollet.
\newblock Depthwise separable convolutions for neural machine translation.
\newblock {\em arXiv preprint arXiv:1706.03059}, 2017.

\bibitem[\protect\citeauthoryear{Li \bgroup \em et al.\egroup
  }{2017}]{li2017mam}
Xuelong Li, Bin Zhao, and Xiaoqiang Lu.
\newblock Mam-rnn: multi-level attention model based rnn for video captioning.
\newblock In {\em Proceedings of the Twenty-Sixth International Joint
  Conference on Artificial Intelligence}, 2017.

\bibitem[\protect\citeauthoryear{Mehri and Sigal}{2018}]{mehri2018middle}
Shikib Mehri and Leonid Sigal.
\newblock Middle-out decoding.
\newblock In {\em Advances in Neural Information Processing Systems}, pages
  5523--5534, 2018.

\bibitem[\protect\citeauthoryear{Oord \bgroup \em et al.\egroup
  }{2016}]{oord2016pixel}
Aaron van~den Oord, Nal Kalchbrenner, and Koray Kavukcuoglu.
\newblock Pixel recurrent neural networks.
\newblock {\em arXiv preprint arXiv:1601.06759}, 2016.

\bibitem[\protect\citeauthoryear{Pan \bgroup \em et al.\egroup
  }{2016a}]{pan2016hierarchical}
Pingbo Pan, Zhongwen Xu, Yi~Yang, Fei Wu, and Yueting Zhuang.
\newblock Hierarchical recurrent neural encoder for video representation with
  application to captioning.
\newblock In {\em Proceedings of the IEEE Conference on Computer Vision and
  Pattern Recognition}, pages 1029--1038, 2016.

\bibitem[\protect\citeauthoryear{Pan \bgroup \em et al.\egroup
  }{2016b}]{pan2016jointly}
Yingwei Pan, Tao Mei, Ting Yao, Houqiang Li, and Yong Rui.
\newblock Jointly modeling embedding and translation to bridge video and
  language.
\newblock In {\em Proceedings of the IEEE conference on computer vision and
  pattern recognition}, pages 4594--4602, 2016.

\bibitem[\protect\citeauthoryear{Papineni \bgroup \em et al.\egroup
  }{2002}]{papineni2002bleu}
Kishore Papineni, Salim Roukos, Todd Ward, and Wei-Jing Zhu.
\newblock Bleu: a method for automatic evaluation of machine translation.
\newblock In {\em Proceedings of the 40th annual meeting on association for
  computational linguistics}, pages 311--318. Association for Computational
  Linguistics, 2002.

\bibitem[\protect\citeauthoryear{Phan \bgroup \em et al.\egroup
  }{2017}]{phan2017consensus}
Sang Phan, Gustav~Eje Henter, Yusuke Miyao, and Shin'ichi Satoh.
\newblock Consensus-based sequence training for video captioning.
\newblock {\em arXiv preprint arXiv:1712.09532}, 2017.

\bibitem[\protect\citeauthoryear{Song \bgroup \em et al.\egroup
  }{2017}]{song2017hierarchical}
Jingkuan Song, Zhao Guo, Lianli Gao, Wu~Liu, Dongxiang Zhang, and Heng~Tao
  Shen.
\newblock Hierarchical lstm with adjusted temporal attention for video
  captioning.
\newblock {\em arXiv preprint arXiv:1706.01231}, 2017.

\bibitem[\protect\citeauthoryear{Sukhbaatar \bgroup \em et al.\egroup
  }{2015}]{sukhbaatar2015end}
Sainbayar Sukhbaatar, Jason Weston, Rob Fergus, et~al.
\newblock End-to-end memory networks.
\newblock In {\em Advances in neural information processing systems}, pages
  2440--2448, 2015.

\bibitem[\protect\citeauthoryear{Sutskever \bgroup \em et al.\egroup
  }{2014}]{sutskever2014sequence}
Ilya Sutskever, Oriol Vinyals, and Quoc~V Le.
\newblock Sequence to sequence learning with neural networks.
\newblock In {\em Advances in neural information processing systems}, pages
  3104--3112, 2014.

\bibitem[\protect\citeauthoryear{Szegedy \bgroup \em et al.\egroup
  }{2015}]{szegedy2015going}
Christian Szegedy, Wei Liu, Yangqing Jia, Pierre Sermanet, Scott Reed, Dragomir
  Anguelov, Dumitru Erhan, Vincent Vanhoucke, and Andrew Rabinovich.
\newblock Going deeper with convolutions.
\newblock In {\em Proceedings of the IEEE conference on computer vision and
  pattern recognition}, pages 1--9, 2015.

\bibitem[\protect\citeauthoryear{Tapaswi \bgroup \em et al.\egroup
  }{2016}]{tapaswi2016movieqa}
Makarand Tapaswi, Yukun Zhu, Rainer Stiefelhagen, Antonio Torralba, Raquel
  Urtasun, and Sanja Fidler.
\newblock Movieqa: Understanding stories in movies through question-answering.
\newblock In {\em Proceedings of the IEEE conference on computer vision and
  pattern recognition}, pages 4631--4640, 2016.

\bibitem[\protect\citeauthoryear{Vaswani \bgroup \em et al.\egroup
  }{2017}]{vaswani2017attention}
Ashish Vaswani, Noam Shazeer, Niki Parmar, Jakob Uszkoreit, Llion Jones,
  Aidan~N Gomez, {\L}ukasz Kaiser, and Illia Polosukhin.
\newblock Attention is all you need.
\newblock In {\em Advances in Neural Information Processing Systems}, pages
  5998--6008, 2017.

\bibitem[\protect\citeauthoryear{Vedantam \bgroup \em et al.\egroup
  }{2015}]{vedantam2015cider}
Ramakrishna Vedantam, C~Lawrence~Zitnick, and Devi Parikh.
\newblock Cider: Consensus-based image description evaluation.
\newblock In {\em Proceedings of the IEEE conference on computer vision and
  pattern recognition}, pages 4566--4575, 2015.

\bibitem[\protect\citeauthoryear{Venugopalan \bgroup \em et al.\egroup
  }{2014}]{venugopalan2014translating}
Subhashini Venugopalan, Huijuan Xu, Jeff Donahue, Marcus Rohrbach, Raymond
  Mooney, and Kate Saenko.
\newblock Translating videos to natural language using deep recurrent neural
  networks.
\newblock {\em arXiv preprint arXiv:1412.4729}, 2014.

\bibitem[\protect\citeauthoryear{Venugopalan \bgroup \em et al.\egroup
  }{2015}]{venugopalan2015sequence}
Subhashini Venugopalan, Marcus Rohrbach, Jeffrey Donahue, Raymond Mooney,
  Trevor Darrell, and Kate Saenko.
\newblock Sequence to sequence-video to text.
\newblock In {\em Proceedings of the IEEE international conference on computer
  vision}, pages 4534--4542, 2015.

\bibitem[\protect\citeauthoryear{Vinyals \bgroup \em et al.\egroup
  }{2015}]{vinyals2015show}
Oriol Vinyals, Alexander Toshev, Samy Bengio, and Dumitru Erhan.
\newblock Show and tell: A neural image caption generator.
\newblock In {\em Proceedings of the IEEE conference on computer vision and
  pattern recognition}, pages 3156--3164, 2015.

\bibitem[\protect\citeauthoryear{Wang \bgroup \em et al.\egroup
  }{2018a}]{wang2018reconstruction}
Bairui Wang, Lin Ma, Wei Zhang, and Wei Liu.
\newblock Reconstruction network for video captioning.
\newblock In {\em Proceedings of the IEEE Conference on Computer Vision and
  Pattern Recognition}, pages 7622--7631, 2018.

\bibitem[\protect\citeauthoryear{Wang \bgroup \em et al.\egroup
  }{2018b}]{wang2018m3}
Junbo Wang, Wei Wang, Yan Huang, Liang Wang, and Tieniu Tan.
\newblock M3: Multimodal memory modelling for video captioning.
\newblock In {\em Proceedings of the IEEE Conference on Computer Vision and
  Pattern Recognition}, pages 7512--7520, 2018.

\bibitem[\protect\citeauthoryear{Weston \bgroup \em et al.\egroup
  }{2014}]{memorynetwork}
Jason Weston, Sumit Chopra, and Antoine Bordes.
\newblock Memory networks.
\newblock {\em arXiv preprint arXiv:1410.3916}, 2014.

\bibitem[\protect\citeauthoryear{Wu and Han}{2018}]{MCF}
Aming Wu and Yahong Han.
\newblock Multi-modal circulant fusion for video-to-language and backward.
\newblock In {\em IJCAI}, pages 1029--1035, 2018.

\bibitem[\protect\citeauthoryear{Xu \bgroup \em et al.\egroup
  }{2016}]{xu2016msr}
Jun Xu, Tao Mei, Ting Yao, and Yong Rui.
\newblock Msr-vtt: A large video description dataset for bridging video and
  language.
\newblock In {\em Proceedings of the IEEE Conference on Computer Vision and
  Pattern Recognition}, pages 5288--5296, 2016.

\bibitem[\protect\citeauthoryear{Xu \bgroup \em et al.\egroup
  }{2017}]{xu2017learning}
Jun Xu, Ting Yao, Yongdong Zhang, and Tao Mei.
\newblock Learning multimodal attention lstm networks for video captioning.
\newblock In {\em Proceedings of the 2017 ACM on Multimedia Conference}, pages
  537--545. ACM, 2017.

\bibitem[\protect\citeauthoryear{Yao \bgroup \em et al.\egroup
  }{2015}]{yao2015describing}
Li~Yao, Atousa Torabi, Kyunghyun Cho, Nicolas Ballas, Christopher Pal, Hugo
  Larochelle, and Aaron Courville.
\newblock Describing videos by exploiting temporal structure.
\newblock In {\em Proceedings of the IEEE international conference on computer
  vision}, pages 4507--4515, 2015.

\bibitem[\protect\citeauthoryear{Yu \bgroup \em et al.\egroup
  }{2016}]{yu2016video}
Haonan Yu, Jiang Wang, Zhiheng Huang, Yi~Yang, and Wei Xu.
\newblock Video paragraph captioning using hierarchical recurrent neural
  networks.
\newblock In {\em Proceedings of the IEEE conference on computer vision and
  pattern recognition}, pages 4584--4593, 2016.

\bibitem[\protect\citeauthoryear{Zhang \bgroup \em et al.\egroup
  }{2016}]{zhang2016augmenting}
Yuting Zhang, Kibok Lee, and Honglak Lee.
\newblock Augmenting supervised neural networks with unsupervised objectives
  for large-scale image classification.
\newblock In {\em International Conference on Machine Learning}, pages
  612--621, 2016.

\end{thebibliography}

\end{document}